# Spatiotemporal Attention Networks for Wind Power Forecasting

Xingbo Fu[1, *], Feng Gao[1], Jiang Wu[1], Xinyu Wei[2], Fangwei Duan[3]

[1]System Engineering Institute, Xi'an Jiaotong University, Xi'an, China

[2]Department of Statistics, Rutgers University, New Brunswick, NJ, USA

[3]Electric Power Research Institute, State Grid Liaoning Electric Power Co. Ltd., Shenyang, China

{xbfu, fgao, jwu}@sei.xjtu.edu.cn, xinyu.wei@rutgers.edu, dfw8906@163.com

*Abstract*—Wind power is one of the most important renewable energy sources and accurate wind power forecasting is very significant for reliable and economic power system operation and control strategies. This paper proposes a novel framework with spatiotemporal attention networks (STAN) for wind power forecasting. This model captures spatial correlations among wind farms and temporal dependencies of wind power time series. First of all, we employ a multi-head self-attention mechanism to extract spatial correlations among wind farms. Then, temporal dependencies are captured by the Sequence-to-Sequence (Seq2Seq) model with a global attention mechanism. Finally, experimental results demonstrate that our model achieves better performance than other baseline approaches. Our work provides useful insights to capture non-Euclidean spatial correlations.

*Index Terms*—Spatiotemporal attention networks, wind power forecasting, attention mechanism.

## I. INTRODUCTION

Wind power is playing a very important role in the electric grid around the world. Due to its variability and stochastic nature, it is difficult to develop a model and predict wind power generation accurately [1]. We need not only to capture temporal dependencies for time series, but also to construct spatial correlations between the target wind farm and some other wind farms.

Wind power forecasting has drawn tremendous attentions from researchers. Some researchers used statistical methods for short-term wind power forecasting. The statistical models include historical average (HA) method and Auto-regressive moving average (ARMA) method [2]. ARMA is the most well-known time-series based method for predicting the future values of wind power and researchers have attempted some variations of ARMA (such as ARIMA) to get better forecasting performance. However, these methods are constrained by the assumption that the target time series is a stationary stochastic process [3]. Unfortunately, wind power generation does not match this assumption in the real world.

Neural networks have been applied widely for time series forecasting. Considering contextual information learning from time series, recurrent neural network (RNN) models can extract explicit temporal dependencies for sequence learning [4]. Moreover, long short-term memory (LSTM) [5] and gated recurrent unit (GRU) [6] are the two special variations of RNN. On the one hand, these approaches have succeeded in many areas including natural language processing (NLP) and time series forecasting. One the other hand, the disadvantage of using these approaches is that they do not adequately consider the spatial dependencies among wind farms.

In neural networks, convolutional neural network (CNN) models are used efficiently to model spatial dependencies for image classification, visual recognition and traffic flow prediction [7]. Nevertheless, CNN specializes in processing data that has a grid-like topology, such as an image [8]. In other words, CNN does not work well when we model the non-Euclidean correlations among different wind farms.

Sequence-to-Sequence (Seq2Seq) is a neural network based on RNN and it has been extensively applied for neural machine translation [9]. The obvious disadvantage of fixed-length context vectors in Seq2Seq models is that it cannot remember the first part of input once it completes processing the whole input. To deal with the incapability of remembering long sequence, attention mechanism was introduced to Seq2Seq models [10]. However, it is hard to parallel in that Seq2Seq models are based on recurrent structures.

Since Transformer was proposed, it has shown impactful capacity to capture temporal dependencies from a sequence by self-attention mechanism. [11] Moreover, Transformer can be paralleled without recurrent network units.

In order to predict wind power generation, we present a novel framework with spatiotemporal attention networks in this paper and the main contributions are as follows:

- Considering correlations of wind farms, we construct spatial self-attention mechanism to capture the spatial dependencies among wind farms;

- To capture time dependencies, temporal attention mechanism is employed with Seq2Seq model;

- Experimental results are presented on real-world wind power datasets, which verify the outperformance of our model comparing with all baselines.

The research presented in this paper is supported in part by the State Grid Science and Technology Program of China.

The rest of this paper is as follows. Firstly, we introduce notations and problem statement in Section II. Next, Section III presents our spatiotemporal attention network for wind power forecasting. In Section IV, we report experimental results of our model in comparison with state-of-the-art baselines. After summarizing related research backgrounds in Section V, we give the conclusions in Section VI.

## II. PRELIMINARIES

### A. Notations

Suppose there are $N$ wind farms, each of which monitors wind power generation time series at that wind farm. Given a time window with $T$ timestamps, $X = (x_1, x_2, \cdots, x_t, \cdots, x_T) \in \mathbb{R}^{N \times T}$ is denoted as wind power generations of all the wind farms for $T$ timestamps. For the $t$th timestamps, we denote $x_t = (x_t^1, x_t^2, \cdots, x_t^N)^T \in \mathbb{R}^{N \times 1}$ as the wind power generation of all the wind farms at timestamp $t$.

### B. Problem Statement

The wind power forecasting problem is to predict the future wind power generations at timestamp $T + n$, which is also a $n$–step ahead prediction. Mathematically, the $n$–step ahead wind power generation $\hat{x}_{T+n}$ can be predicted by

$$\hat{x}_{T+n} = f(X) = f(x_1, x_2, \cdots, x_t, \cdots, x_T) \quad (1)$$

## III. SPATIOTEMPORAL ATTENTION NETWORKS

In this section, we introduce the structure of spatiotemporal attention networks (STAN). Fig. 1 illustrates the architecture of our proposed model.

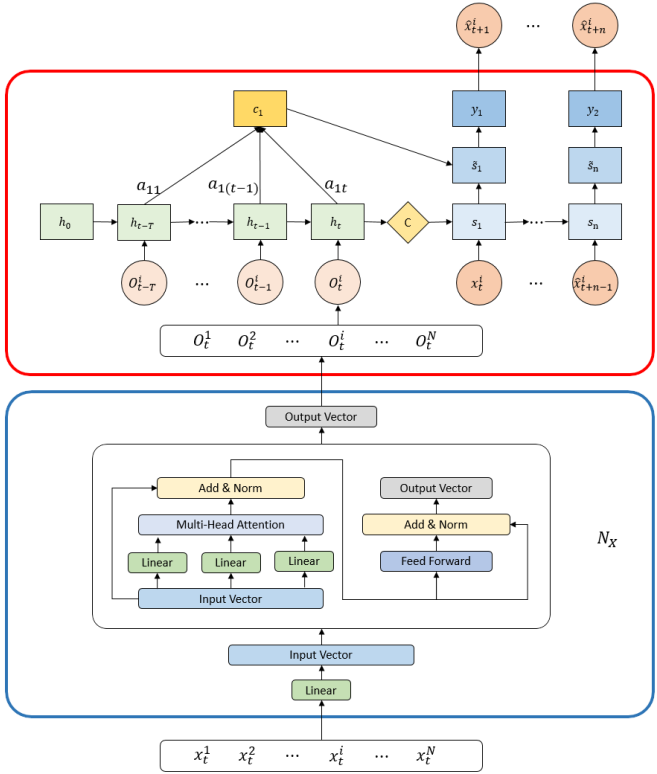

Fig. 1 Architecture of STAN. STAN model consists of spatial self-attention mechanism (blue box) and temporal attention mechanism (red box).

### A. Spatial self-attention mechanism

Following Transformer [11], a spatial self-attention mechanism employed in this model is used to extract spatial correlations among wind farms. The input of this module is the input vector $I_t$ encoded by a simple fully connected feed-forward network from the wind power generation of all the wind farms at timestamp $t$.

$$I_t = x_t W^I \quad (2)$$

where $W^I \in \mathbb{R}^{1 \times d_m}$ is the learnable weight matrix and $d_m$ is the dimension of this model.

Once getting input vector, it is processed by a multi-head attention and a fully connected feed-forward network.

#### 1) Multi-Head Attention

The structure of multi-head attention is shown as Fig. 2.

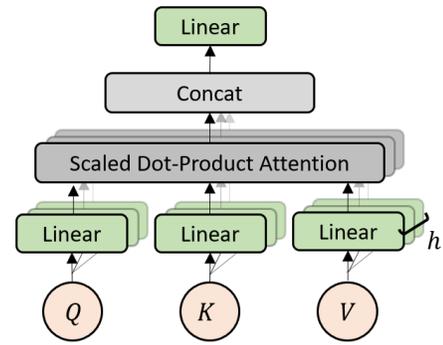

Fig. 2 Architecture of Multi-Head Attention.

Multi-head attention contains a query vector $Q \in \mathbb{R}^{N \times d_m}$, a key vector $K \in \mathbb{R}^{N \times d_m}$ and a value vector $V \in \mathbb{R}^{N \times d_m}$, all of which are the input vector $I_t \in \mathbb{R}^{N \times d_m}$ at timestamp $t$. From these three vectors, we compute scaled dot-product attention [11] function to get the output as follows:

$$Attention(Q, K, V) = \text{softmax}\left(\frac{QK^T}{\sqrt{d_m}}\right) V \quad (3)$$

As (3) shows, the output is a weight sum of the values and the weight assigned to each value is calculated by the dot-product of the query with all keys. This scaled dot-product attention function is different from Luong attention mechanism because of the introduction of a scaling factor $\frac{1}{\sqrt{d_m}}$. When the dimension $d_m$ is a large value, the dot product significantly fluctuates. To offset this effect, the dot product is scaled by the scaling factor $\frac{1}{\sqrt{d_m}}$.

As for "multi-head", each of these three vectors $Q, K, V$ is linearly projected $h$ times respectively as follows:

$$Q_i' = Q W_i^Q, \quad i = 1, 2, \cdots, h \quad (4)$$

$$K_i' = K W_i^K, \quad i = 1, 2, \cdots, h \quad (5)$$

$$W_i' = V W_i^V, \quad i = 1, 2, \cdots, h \quad (6)$$

where $W_i^Q \in \mathbb{R}^{d_m \times d_k}$, $W_i^K \in \mathbb{R}^{d_m \times d_k}$ and $W_i^V \in \mathbb{R}^{d_m \times d_k}$ are the learnable weight matrices and $d_k = d_m/h$.

Therefore, each attention function is computed as follows:

$$head_i = Attehtion(Q'_i, K'_i, V'_i) \quad (7)$$

These attention function are concatenated and projected resulting in the output of multi-head attention as follows:

$$\begin{aligned}MultiHead(Q,K,V) \\ = Concat(head_1, head_2, \cdots, head_h)W^O\end{aligned} \quad (8)$$

where $W^O \in \mathbb{R}^{d_m \times d_m}$ is the learnable weight matrix.

*2) Feed-Forward Networks*

After multi-head attention, we employ a fully connected feed-forward network. This network contains of two linear transformations with a ReLU activation between them.

$$FFN(x) = [ReLU(xW^1)]W^2 \quad (9)$$

where $W^1 \in \mathbb{R}^{d_m \times d_{ffn}}$ and $W^2 \in \mathbb{R}^{d_{ffn} \times d_m}$ are the learnable weight matrices and $d_{ffn}$ is the size of the inner-layer.

*3) Residual Connections and Layer Normalizations*

We take the output vector as input vector and feed it back to multi-head attention. Repeating this processing $N_X$ times, the model constructs a deep $k$-layer spatial self-attention.

To conquer the degradation problem with deeper networks [12], residual connections are employed after each multi-head attention and feed-forward network. Following residual connections, layer normalizations restrict weights to a certain range [13]. Residual connections and layer normalizations get the model to train effectively.

Residual connections and layer normalizations following multi-head attention can be written as follows:

$$ATTN_t = LN[I_t + MultiHead(Q,K,V)] \quad (10)$$

where $LN$ is the layer normalization function.

Residual connections and layer normalizations following the feed-forward network can be written as follows:

$$O_t = LN[ATTN_t + FFN(ATTN_t)] \quad (11)$$

where $O_t \in \mathbb{R}^{N \times d_m}$ and $LN$ is the layer normalization function.

Therefore, we obtain the output of spatial self-attention mechanism.

*B. Temporal attention mechanism*

Temporal attention mechanism in this model is a Seq2Seq model based on RNN with global attention mechanism [14].

The Seq2Seq model consists of an encoder and a decoder. The encoder's input is $O_t^i \in \mathbb{R}^{1 \times d_m}$, the corresponding part to target wind farm of $O_t$ from timestamp 1 to $T$.

At timestamp $t$, the hidden state $h_t$ in the encoder is computed as follows:

$$h_t = \varphi(O_t^i U^E + h_{t-1} W^E) \quad (12)$$

Where $U^E \in \mathbb{R}^{d_m \times d_e}$ and $W^E \in \mathbb{R}^{d_e \times d_e}$ are the learnable weight matrices in the encoder and $d_e$ is the dimension of this model. $\varphi$ is a tanh activation. $h_0$ is the zero state.

The context vector $c_i$ captures relevant information from the encoder to predict the future values. The difference between Seq2Seq models and Seq2Seq models with attention mechanism is that the attention mechanism dynamically computes the context vector for each timestamp. $c_k$ is computed as the weighted average over all the hidden states of the encoder as follows:

$$c_k = \sum_{j=1}^{T} a_{kj} h_j \quad (13)$$

where $a_{kj}$ is an element of weight vector $a_t \in \mathbb{R}^{1 \times T}$ and $a_{kj}$ is computed as follows:

$$a_{kj} = \frac{exp\left(score(s_k, h_j)\right)}{\sum_{j=1}^{T} exp\left(score(s_k, h_j)\right)} \quad (14)$$

where $s_k \in \mathbb{R}^{d_d \times 1}$ is the hidden state of the decoder and $d_d$ is the size of hidden cells in the decoder.

As (14) shows, $a_{kj}$ is a softmax result of a score function. According to Luong attention mechanism [14], the score function has three different alternatives. In this paper, we implement the general score function as follows:

$$score(s_k, h_j) = s_k^T W^I h_j \quad (15)$$

where $W^I \in \mathbb{R}^{d_d \times d_e}$ is the learnable weight matrix.

Given the hidden state $s_i$ of the decoder and the context vector $c_i$, a simple concatenation is used to combine the information from both vectors to produce an attentional hidden state as follows:

$$\tilde{s}_k = tanh([c_k; s_k]W^C) \quad (16)$$

where $[;]$ is a concentration operation and $W^C \in \mathbb{R}^{(d_e+d_d) \times d_d}$ is the learnable weight matrix.

At last, we employ a linear projection to make the output of decoder as follows:

$$y_k = \tilde{s}_k W^S \quad (17)$$

where $W^S \in \mathbb{R}^{d_d \times 1}$ is the learnable weight matrix.

Finally, the prediction $\hat{x}_{T+k}^i$ is

$$\hat{x}_{T+k}^i = y_k \quad (18)$$

IV. EXPERIMENTS

Experiments are conducted on real-world datasets to evaluate the performance of our proposed spatiotemporal attention networks. In this section, we introduce the datasets and baseline algorithms used along with experimental setup and the evaluation. The code for our proposed model is available on https://github.com/xbfu/Spatiotemporal-Attention-Networks.

*A. Datasets*

We use the wind power generation dataset collected by National Renewable Energy Laboratory (NREL). This dataset records wind power generation of 1325 wind farms with a ten-minute sampling rate from 2004 to 2006. We select 6 wind farms among the dataset. Table I shows the locations of these six wind farms.

B. Experimental Settings

- Loss Function: During the training, we use Adam optimizer [2014] to train the model by minimizing the mean squared error (MSE) between the prediction and the ground truth.

$$MSE = \frac{1}{N}\sum_{i=1}^{N}(x_{t+n}^i - \hat{x}_{t+n}^i)^2 \quad (19)$$

- Evaluation Metrics: To measure the effectiveness of our proposed model and baseline algorithms, we use the root mean square error (RMSE) as the evaluation metrics.

$$RMSE = \sqrt{\frac{1}{N}\sum_{i=1}^{N}(x_{t+n}^i - \hat{x}_{t+n}^i)^2} \quad (20)$$

- Hyperparameters: The hyperparameters of our model are shown in Table II.

C. Baseline Algorithms

We compare our proposed model with the following baseline algorithms.

- HA: Historical Average uses the average of previous observations as the prediction.
- ARIMA: A variation of ARMA and one of the most widely used methods for time series prediction.
- ANN [16]: Artificial Neural Network is also widely applied in time series forecasting. In this paper, we construct an ANN with a single hidden layer which has 100 hidden units.
- GRU [6]: Gated recurrent units are a gating mechanism in recurrent neural networks. In this paper, we construct two GRU models: GRUs with the input of target win farm and GRUm with the input of all the wind farms.
- Seq2Seq [9]: Seq2Seq models consist of an encoder and a Decoder. The encoder maps input to a fixed-length context vector and the decoder generates output according to the context vector.
- Seq2SeqAttn [10]: Seq2Seq models with global attention mechanism does not assume a monotonic alignment, which is different from Seq2Seq.

Furthermore, to fully evaluate the performance benefiting from each component of our proposed approach, we implement two degraded versions of STAN.

- STANsa: This variation of STAN consists of spatial self-attention mechanism and Seq2Seq model. In other words, we remove the temporal attention mechanism.
- STANta: We replace spatial self-attention mechanism with a simple fully connected feed-forward network. The difference between STANta and Seq2SeqAttn is that the input of Seq2SeqAttn is only the target wind farm.

D. Experimental Results

The following experiments are performed to evaluate our proposed approach. This part contains two evaluations: accuracy comparison and converge speed comparison.

*1) Accuracy Comparison*

We conduct experiments on wind power generation datasets to compare the performance of our proposed model with its two variations and seven baseline algorithms. Table III shows accuracy comparisons for different steps among these ten approaches. As suggested in Table III, the STAN outperforms seven baseline approaches.

Table I Locations of 6 Wind Farms

| ID | Latitude | Longitude |
|---|---|---|
| SITE_00173 | 36.14 | -100.34 |
| SITE_00193 | 36.42 | -100.44 |
| SITE_00215 | 36.42 | -100.67 |
| SITE_00365 | 36.50 | -100.68 |
| SITE_00446 | 36.50 | -100.28 |
| SITE_00797 | 36.56 | -100.54 |

Table II Hyperparameters

| Parameters | Values |
|---|---|
| $d_m, d_d, d_e$ | 512 |
| $T$ | 12 |
| $h$ | 8 |
| $d_{ffn}$ | 2048 |
| $N_X$ | 6 |
| Training Epochs | 40 |
| Learning Rate | 0.01 |

Table III Accuracy Comparison of Different Models

| NO. | Method | RMSE | | |
|---|---|---|---|---|
| | | 1-step | 2-step | 3step |
| 1 | HA | 54.54 | 72.32 | 91.74 |
| 2 | ARIMA | 35.77 | 67.03 | 97.91 |
| 3 | ANN | 31.58 | 61.15 | 96.38 |
| 4 | GRUs | 31.40 | 58.79 | 77.36 |
| 5 | GRUm | 29.14 | 58.31 | 75.22 |
| 6 | Seq2Seq | 30.21 | 62.41 | 86.4 |
| 7 | Seq2SeqAttn | 27.72 | 63.28 | 81.60 |
| 8 | STANsa | 28.89 | 58.39 | 74.23 |
| 9 | STANta | 27.29 | 58.41 | 75.91 |
| 10 | STAN | 25.82 | 57.22 | 73.67 |

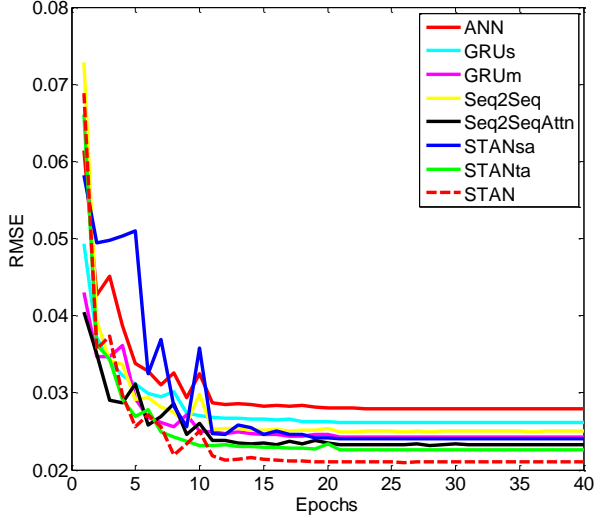

Fig. 3 Converging Speed of Models

*2) Converging Speed Comparison*

During the training, rapidly convergence is important. This evaluation investigates the converging speeds for 1-step forecasting of different models. The result is shown in Fig. 3 and it suggests that the STAN model converges faster than other baseline approaches.

## V. RELATED WORK

### A. Wind Power Forecasting

Wind power forecasting has two common approaches. The first is the physical model using parameterizations based on a detailed physical description of the atmosphere around and insider the wind farm. Numeric Weather Prediction (NWP) is the most well-known method [17]. On the one hand, NWP is very reliable since it predicts wind speed by solving complicated mathematical models with weather data like temperature, surface roughness and obstacles. Then we can extrapolate wind power according to the relationship between wind speed and output. On the other hand, the temporal resolution is usually between 1 and 3 hours in NWP and sometimes we can only get 4-time predictions at specific nodes of a grid per day [2]. Therefore, NWP model is an ideal method for long-term wind power forecasting but not for short-term.

The second model is statistical approach based on historical data of power. Sub-classification of statistical is time-series based models and neural network based models. One of the time-series based models is ARMA with its variations ARIMA and ARIMAX (ARMA with exogenous input). These linear predictors with the assumption that time series is stationary stochastic process. Neural network based models, or artificial-intelligence-based models construct the relationship between historical data and output. Neural networks contain fully connected network (FCN), stacked autoencoder (SAE) [18], convolutional neural network (CNN), recurrent neural network (RNN), etc.

The challenge of wind power forecasting is modeling spatial correlations and temporal correlations. Temporal correlations in wind power time series are obvious while spatial correlations are also indispensable.

### B. Self-Attention Mechanism

Self-attention is an attention mechanism relating different positions of a sequence and captures correlations among each part of this sequence. Recently it has been widely applied in neural machine translation (NMT), abstractive summarization, etc.

Transformer is one of the most impressive frameworks based on the self-attention mechanisms [11]. This work is totally different from the past sequence model such as RNN and Seq2Seq model.

In Transformer, "Scaled Dot-Product Attention" is firstly employed to capture correlations among each input of sequence by attention function. Attention weights are calculated by the dot products of the query with all keys, divided by $\sqrt{d_k}$ ($d_k$ is the dimension of queries and keys) and a softmax function on the values.

The other pioneering structure of Transformer is the multi-headed attention mechanism. Transformer projects the queries, keys and values $h$ times to get $h$ attention functions instead of performing a single attention function. Then these projected attention functions are concatenated leading to the final values. This module makes it possible for attention mechanism to concentrate different parts of input.

In this paper, we leverage self-attention mechanism to capture spatial dependencies among different wind farms inspired by Transformer.

### C. Seq2Seq with Attention

Based on the Seq2Seq model, Seq2Seq model with attention uses dynamic context vector which are computed according to all the output of the encoder. This attention mechanism can focus partially on input sequence to accurately remember and process long complex temporal dependencies.

The context vector, which is used to calculate the final output of Decoder, is calculated by the dot product of weight vector and the output of the encoder. As for weight vector, we need to employ a softmax function on score function. Luong attention mechanism proposed three types of score function [14]. In this paper, we use the general score function.

## VI. CONCLUSION

In this paper, we propose a novel framework with spatiotemporal attention networks for wind power forecasting. This model contains a spatial self-attention mechanism used to extract spatial correlations among different wind farms along with a temporal attention mechanism to capture temporal dependencies. These two attention mechanism efficiently model spatiotemporal problems with non-Euclidean correlations. The experiment is conducted on real-world datasets and our proposed model delivers the best performance over seven baseline algorithms. In the future work, we will focus on intersections between graph convolutional networks and attention mechanisms with their applications to spatiotemporal modeling.

## REFERENCES

[1] Zhao Y, Ye L, Pinson P, et al. Correlation-constrained and sparsity-controlled vector autoregressive model for spatio-temporal wind power forecasting[J]. IEEE Transactions on Power Systems, 2018, 33(5): 5029-5040.

[2] Soman, Saurabh S., et al. "A review of wind power and wind speed forecasting methods with different time horizons." North American Power Symposium 2010. IEEE, 2010.

[3] Box, George EP, and David A. Pierce. "Distribution of residual autocorrelations in autoregressive-integrated moving average time series models." Journal of the American Statistical Association 65.332 (1970): 1509-1526.

[4] Kariniotakis, G. N., G. S. Stavrakakis, and E. F. Nogaret. "Wind power forecasting using advanced neural networks models." IEEE transactions on Energy conversion 11.4 (1996): 762-767.

[5] Hochreiter, Sepp, and Jürgen Schmidhuber. "Long short-term memory." Neural computation 9.8 (1997): 1735-1780.

[6] Cho K, Van Merriënboer B, Gulcehre C, et al. Learning phrase representations using RNN encoder-decoder for statistical machine translation[J]. arXiv preprint arXiv:1406.1078, 2014.

[7] Zhang, Junbo, et al. "Predicting citywide crowd flows using deep spatio-temporal residual networks." Artificial Intelligence 259 (2018): 147-166.

[8] Geng, Xu, et al. "Spatiotemporal multi-graph convolution network for ride-hailing demand forecasting." 2019 AAAI Conference on Artificial Intelligence (AAAI'19). 2019.

[9] Sutskever, Ilya, Oriol Vinyals, and Quoc V. Le. "Sequence to sequence learning with neural networks." Advances in neural information processing systems. 2014.

[10] Bahdanau, Dzmitry, Kyunghyun Cho, and Yoshua Bengio. "Neural machine translation by jointly learning to align and translate." arXiv preprint arXiv:1409.0473 (2014).

[11] Vaswani, Ashish, et al. "Attention is all you need." Advances in neural information processing systems. 2017.

[12] He, Kaiming, et al. "Deep residual learning for image recognition." Proceedings of the IEEE conference on computer vision and pattern recognition. 2016.

[13] Ba, Jimmy Lei, Jamie Ryan Kiros, and Geoffrey E. Hinton. "Layer normalization." arXiv preprint arXiv:1607.06450 (2016).

[14] Luong, Minh-Thang, Hieu Pham, and Christopher D. Manning. "Effective approaches to attention-based neural machine translation." arXiv preprint arXiv:1508.04025 (2015).

[15] https://www.nrel.gov/

[16] Wang, Lin, Yi Zeng, and Tao Chen. "Back propagation neural network with adaptive differential evolution algorithm for time series forecasting." Expert Systems with Applications 42.2 (2015): 855-863.

[17] Higashiyama, Kazutoshi, Yu Fujimoto, and Yasuhiro Hayashi. "Feature Extraction of NWP Data for Wind Power Forecasting Using 3D-Convolutional Neural Networks." Energy Procedia 155 (2018): 350-358.

[18] Li X, Peng L, Hu Y, et al. Deep learning architecture for air quality predictions[J]. Environmental Science and Pollution Research, 2016, 23(22):22408-22417.

[19] Jiang Z, Shekhar S, Zhou X, et al. Focal-Test-Based Spatial Decision Tree Learning[J]. IEEE Transactions on Knowledge and Data Engineering, 2015, 27(6):1547-1559.

[20] Yuan Z, Zhou X, Yang T. Hetero-convlstm: A deep learning approach to traffic accident prediction on heterogeneous spatio-temporal data[C]//Proceedings of the 24th ACM SIGKDD International Conference on Knowledge Discovery & Data Mining. ACM, 2018: 984-992.